\documentclass[11pt]{article}

\usepackage[utf8]{inputenc}
\usepackage[T1]{fontenc}
\usepackage{lmodern}
\usepackage{amsmath,amssymb,amsthm}
\usepackage{graphicx}
\usepackage{float}
\usepackage{geometry}
 \usepackage[round]{natbib}   

\usepackage{algorithm}
\usepackage{algpseudocode}   

\usepackage{hyperref}

\theoremstyle{plain}

\theoremstyle{definition}

\theoremstyle{remark}

\title{Weighted Support Points from Random Measures:\\ An Interpretable Alternative for Generative Modeling}
\author{
  Peiqi Zhao \\
  Department of Mathematics and Statistics and Data Sciences, \\
  University of Texas at Austin \\
  \and
  Carlos E. Rodríguez\thanks{Corresponding author: \href{mailto:carloserwin@sigma.iimas.unam.mx}{carloserwin@sigma.iimas.unam.mx}} \\
  Department of Probability and Statistics, \\
  IIMAS, Universidad Nacional Autónoma de México \\
  \and
  Ramsés H. Mena \\
  Department of Probability and Statistics, \\
  IIMAS, Universidad Nacional Autónoma de México \\
  \and
  Stephen G. Walker \\
  Department of Mathematics and Statistics and Data Sciences, \\
  University of Texas at Austin
}
\date{\today}

\begin{document}

\maketitle

\begin{abstract}
Support points summarize a large dataset through a smaller set of representative points that can be used for data operations, such as Monte Carlo integration, without requiring access to the full dataset. In this sense, support points offer a compact yet informative representation of the original data. We build on this idea to introduce a generative modeling framework based on random weighted support points, where the randomness arises from a weighting scheme inspired by the Dirichlet process and the Bayesian bootstrap. The proposed method generates diverse and interpretable sample sets from a fixed dataset, without relying on probabilistic modeling assumptions or neural network architectures. We present the theoretical formulation of the method and develop an efficient optimization algorithm based on the Convex--Concave Procedure (CCP). Empirical results on the MNIST and CelebA-HQ datasets show that our approach produces high-quality and diverse outputs at a fraction of the computational cost of black-box alternatives such as Generative Adversarial Networks (GANs) or Denoising Diffusion Probabilistic Models (DDPMs). These results suggest that random weighted support points offer a principled, scalable, and interpretable alternative for generative modeling. A key feature is their ability to produce genuinely interpolative samples that preserve underlying data structure.
\end{abstract}


\section{Introduction}


Generative models are the cornerstone of modern machine learning. Their goal is to capture the underlying probability distribution of observed data in order to generate new samples that resemble the original dataset \citep{ho2020denoising, kingma2013auto}. This ability to create realistic and diverse data, particularly in complex domains such as images, has opened up transformative applications in computer vision, digital art, data augmentation, and scientific simulation \citep{karras2019style, rombach2022high, dhariwal2021diffusion, brock2018large, karras2020analyzing}. Advances in deep learning have played a central role in this progress, enabling models to learn intricate data structures that were previously out of reach.

Two of the most prominent families of deep generative models are Generative Adversarial Networks (GANs) \citep{goodfellow2014generative, radford2015unsupervised} and Denoising Diffusion Probabilistic Models (DDPMs) \citep{ho2020denoising, sohl2015deep, song2019generative}. GANs are based on a game-theoretic setup, where a generator learns to produce realistic samples while a discriminator tries to distinguish them from real data \citep{goodfellow2014generative}. In contrast, DDPMs are inspired by physical diffusion processes and rely on a two-stage procedure: a forward process that gradually corrupts the data by adding noise, and a learned reverse process that reconstructs samples from noise \citep{ho2020denoising, sohl2015deep}. Both approaches have achieved remarkable results in image synthesis, producing high-resolution and visually compelling samples in a variety of applications \citep{ho2020denoising, karras2019style}.


Despite their impressive performance, established generative models like GANs and DDPMs face important limitations. At the same time, they share a reliance on deep neural networks to learn complex data distributions, while differing significantly in their architectures, training procedures, and computational trade-offs.

GANs consist of two neural networks, a generator and a discriminator, that are trained simultaneously in an adversarial setting. The generator learns to produce realistic samples, while the discriminator tries to distinguish real from synthetic data \citep{goodfellow2014generative}. Although GANs can generate high-quality images efficiently once trained, the training process is notoriously unstable. It often suffers from issues such as mode collapse, where the generator produces a narrow set of outputs, and requires a delicate balance between the two networks to converge successfully \citep{goodfellow2016nips, salimans2016improved}.

In contrast, DDPMs typically involve a single, large neural network, often based on a U-Net architecture with millions of parameters, trained to reverse a noise corruption process. Generating a single sample requires hundreds or even thousands of iterative denoising steps, each invoking the network. This makes both training and inference computationally intensive. Moreover, the multi-step generative process complicates interpretation and renders the model a black box, despite its strong empirical performance \citep{nichol2021improved, song2020denoising}.


These limitations encourage exploring alternative generative frameworks that provide simpler training, reduced computational costs, or fresh theoretical perspectives for producing high-quality data.


This work explores the use of \emph{support points} for generative modeling, proposing a mathematically grounded, nonparametric alternative to black-box neural methods such as GANs and diffusion models. Originally introduced by \citet{mak2018support}, support points are defined as a set of representative locations that minimize the energy distance between a target distribution \( F \) and its discrete approximation \( \widehat{F}_n \) for a chosen sample size $n$ \citep{szekely2004testing, szekely2013energy}. The energy distance is a geometric divergence widely used in nonparametric hypothesis testing and distributional approximation. While support points have proven effective for tasks such as distribution compression, uncertainty quantification, and MCMC summarization \citep{mak2018support, riachy2019stein, mak2017projected, joseph2015sequential, joseph2019deterministic}, their potential for generative modeling, particularly in high-dimensional settings, remains un-explored.

The central contribution of this work is a framework for \emph{weighted support points} driven by a \emph{finite random measure} designed to balance representativeness and diversity. Instead of reweighting all \(N\) observations under Dirichlet–process or Bayesian–bootstrap schemes \citep{ferguson1973,rubin1981bayesian}, we construct a reweighted empirical measure by first selecting a random subset of the reference data and then assigning exchangeable weights from a symmetric Dirichlet distribution. The resulting measure remains centered at the empirical distribution and has \emph{tunable dispersion}, encouraging genuine run-to-run variability without collapsing mass onto a few atoms.

Given this random measure, we compute support points of size \(n\) by minimizing the energy distance between it and the empirical distribution of the candidate set. This leads to a stochastic optimization whose solutions differ across realizations, yielding diverse yet data-aligned configurations from the same reference dataset. The subsetting rule and the calibration of weight dispersion are detailed in the body of the paper. Empirical results on MNIST and CelebA-HQ datasets, including comparisons with GANs and DDPMs, show that the proposed method produces high-quality, diverse outputs at a fraction of the computational cost, offering a principled, interpretable, and scalable alternative for generative modeling.


\section{Background: Established Generative Models}

This section provides a mathematical overview of two prominent generative modeling frameworks: Generative Adversarial Networks (GANs) and Denoising Diffusion Probabilistic Models (DDPMs). Understanding their core principles is essential for contextualizing and comparing the support points approach.

\subsection{Generative Adversarial Networks (GANs)}

GANs, introduced by \citet{goodfellow2014generative}, represent a milestone in generative modeling by casting the learning process as a two-player minimax game between a generator and a discriminator.

Training unfolds as a competitive interaction: the generator attempts to fool the discriminator, while the discriminator aims to correctly distinguish real from synthetic samples. The generator, denoted \( G(\mathbf{z}; \theta_g) \), maps latent noise vectors \( \mathbf{z} \sim p_z(\mathbf{z}) \) (typically sampled from a standard normal or uniform distribution) to synthetic data samples \( \mathbf{x}_g = G(\mathbf{z}) \). The discriminator, \( D(\mathbf{x}; \theta_d) \), receives a sample \( \mathbf{x} \) and outputs the estimated probability that \( \mathbf{x} \) originates from the real data distribution \( p_{\text{data}} \), rather than from the generator’s distribution \( p_g \).

The networks are trained by solving the following minimax optimization problem:
\begin{align}
\min_G \max_D V(D, G) &= \mathbb{E}_{\mathbf{x} \sim p_{\text{data}}}[\log D(\mathbf{x})] \nonumber\\ &\quad + \mathbb{E}_{\mathbf{z} \sim p_z}[\log(1 - D(G(\mathbf{z})))].
\label{eq:gan_minimax}
\end{align}

This adversarial setup incentivizes the generator to produce increasingly realistic data. At equilibrium, the generator perfectly replicates the true data distribution (\( p_g = p_{\text{data}} \)), and the discriminator is maximally uncertain, yielding \( D(\mathbf{x}) = 1/2 \) for all \( \mathbf{x} \).

For a fixed generator \( G \), the optimal discriminator is
\begin{equation*}
D^*_G(\mathbf{x}) = \frac{p_{\text{data}}(\mathbf{x})}{p_{\text{data}}(\mathbf{x}) + p_g(\mathbf{x})},
\end{equation*}
which, when substituted into the value function \( V(D, G) \), leads to an objective for the generator equivalent to minimizing the Jensen--Shannon divergence:
\begin{equation*}
\min_G V(D^*_G, G) = -\log 4 + 2 \cdot \text{JSD}(p_{\text{data}} \,\|\, p_g).
\end{equation*}

In practice, training GANs can be challenging. The original loss in Equation~\eqref{eq:gan_minimax} often results in vanishing gradients for the generator, particularly when \( D(G(\mathbf{z})) \approx 0 \). To address this, a widely adopted modification is to train the generator using the non-saturating heuristic:
\[
\max_G \mathbb{E}_{\mathbf{z} \sim p_z} [\log D(G(\mathbf{z}))],
\]
which yields stronger and more informative gradients during early stages of training.

GANs have been successfully applied in tasks such as image generation, domain translation, and data augmentation. Numerous extensions address training instability, including Deep Convolutional GANs (DCGAN) \citep{radford2015unsupervised}, Wasserstein GANs (WGAN) \citep{arjovsky2017wasserstein}, among others.

\subsection{Denoising Diffusion Probabilistic Models (DDPMs)}

DDPMs have become a leading generative modeling technique, achieving remarkable performance in image, audio, and multimodal generation tasks \citep{ho2020denoising, nichol2021improved, dhariwal2021diffusion}. Their central idea is conceptually simple but mathematically rich: progressively corrupt the training data by adding Gaussian noise, and then learn to reverse this process step by step to generate new data.

The diffusion process is structured as a pair of Markov chains: a fixed \emph{forward process} that adds noise to data, and a learned \emph{reverse process} that denoises this input back to the data distribution.

\subsubsection{Forward Process}

Let \( \boldsymbol{x}_0 \sim q(\boldsymbol{x}_0) \) be a data point. The forward process generates a sequence \( \boldsymbol{x}_1, \dots, \boldsymbol{x}_T \) by recursively adding Gaussian noise:
\begin{equation}
    \boldsymbol{x}_t = \sqrt{1 - \beta_t} \, \boldsymbol{x}_{t-1} + \sqrt{\beta_t} \, \boldsymbol{\epsilon}_t, \quad \boldsymbol{\epsilon}_t \sim \mathcal{N}(\boldsymbol{0}, \boldsymbol{I}), \quad t = 1, \dots, T,
    \label{eq:ddpm_forward}
\end{equation}
where \( \{ \beta_t \}_{t=1}^T \) is a predefined variance schedule. Typically, \( \beta_t \) increases linearly from small values (e.g., \( 10^{-4} \)) to moderate values (e.g., \( 0.02 \)) \citep{ho2020denoising}.

A key result is that \( \boldsymbol{x}_t \) can be sampled directly from \( \boldsymbol{x}_0 \) without computing intermediate steps:
\begin{equation}
    \boldsymbol{x}_t = \sqrt{\bar{\alpha}_t} \, \boldsymbol{x}_0 + \sqrt{1 - \bar{\alpha}_t} \, \boldsymbol{\epsilon}, \quad \boldsymbol{\epsilon} \sim \mathcal{N}(\boldsymbol{0}, \boldsymbol{I}),
    \label{eq:ddpm_closed}
\end{equation}
where \( \alpha_t = 1 - \beta_t \) and \( \bar{\alpha}_t = \prod_{s=1}^t \alpha_s \).

\subsubsection{Reverse Process}

At test time, generation begins with \( \boldsymbol{x}_T \sim \mathcal{N}(\boldsymbol{0}, \boldsymbol{I}) \), and a learned reverse process approximates the true posterior \( q(\boldsymbol{x}_{t-1} | \boldsymbol{x}_t) \). Each transition is modeled as a Gaussian:
\[
p_\theta(\boldsymbol{x}_{t-1} | \boldsymbol{x}_t) = \mathcal{N}(\boldsymbol{x}_{t-1}; \boldsymbol{\mu}_\theta(\boldsymbol{x}_t, t), \sigma_t^2 \boldsymbol{I}),
\]
where \( \boldsymbol{\mu}_\theta \) is predicted by a neural network, often a U-Net \citep{ronneberger2015u, ho2020denoising}. The variance \( \sigma_t^2 \) is fixed or learned, and commonly chosen as \( \beta_t \) or \( \tilde{\beta}_t = \frac{1 - \bar{\alpha}_{t-1}}{1 - \bar{\alpha}_t} \beta_t \) \citep{nichol2021improved}.

Using Bayes' rule, we can write in closed form:
\[
q(\boldsymbol{x}_{t-1} | \boldsymbol{x}_t, \boldsymbol{x}_0) = \mathcal{N}(\boldsymbol{x}_{t-1}; \tilde{\boldsymbol{\mu}}_t, \tilde{\beta}_t \mathbf{I}),
\]
with
\[
\tilde{\boldsymbol{\mu}}_t(\boldsymbol{x}_t, \boldsymbol{x}_0) = \frac{\sqrt{\bar{\alpha}_{t-1}} \beta_t}{1 - \bar{\alpha}_t} \, \boldsymbol{x}_0 + \frac{\sqrt{\alpha_t} (1 - \bar{\alpha}_{t-1})}{1 - \bar{\alpha}_t} \, \boldsymbol{x}_t.
\]
Since \( \boldsymbol{x}_0 \) is not available at test time, the network learns to predict either \( \boldsymbol{x}_0 \) or, more commonly, the noise \( \boldsymbol{\epsilon} \) used in the forward process. Using Eq.~\eqref{eq:ddpm_closed}, we write:
\[
\boldsymbol{x}_0 = \frac{1}{\sqrt{\bar{\alpha}_t}} \left( \boldsymbol{x}_t - \sqrt{1 - \bar{\alpha}_t} \, \boldsymbol{\epsilon} \right),
\]
so the model can instead predict \( \boldsymbol{\epsilon}_\theta(\boldsymbol{x}_t, t) \) and compute:
\begin{equation}
    \boldsymbol{\mu}_\theta(\boldsymbol{x}_t, t) = \frac{1}{\sqrt{\alpha_t}} \left( \boldsymbol{x}_t - \frac{\beta_t}{\sqrt{1 - \bar{\alpha}_t}} \boldsymbol{\epsilon}_\theta(\boldsymbol{x}_t, t) \right).
    \label{eq:ddpm_reverse_mean}
\end{equation}

\subsubsection{Training and Sampling}

The model is trained by minimizing a simplified version of the variational lower bound:
\begin{equation}
L_{\text{simple}}(\theta) = \mathbb{E}_{t, \boldsymbol{x}_0, \boldsymbol{\epsilon}} \left[ \left\| \boldsymbol{\epsilon} - \boldsymbol{\epsilon}_\theta(\boldsymbol{x}_t, t) \right\|^2 \right],
\label{eq:ddpm_simple_objective}
\end{equation}
where \( \boldsymbol{x}_t \) is sampled via Eq.~\eqref{eq:ddpm_closed}. This trains the network to recover the added noise at any given timestep. At inference time, the generation algorithm applies Eq.~\eqref{eq:ddpm_reverse_mean} recursively, starting from \( \boldsymbol{x}_T \sim \mathcal{N}(\mathbf{0}, \mathbf{I}) \) and sampling \( \boldsymbol{x}_{t-1} \) from:
\[
\boldsymbol{x}_{t-1} = \boldsymbol{\mu}_\theta(\boldsymbol{x}_t, t) + \sigma_t \boldsymbol{z}, \quad \boldsymbol{z} \sim \mathcal{N}(\boldsymbol{0}, \mathbf{I}).
\]

Foundational papers and tutorials include \citep{ho2020denoising, sohl2015deep, song2019generative, nichol2021improved, song2020denoising, dhariwal2021diffusion, ronneberger2015u, kingma2019introduction, turner2024ddpm_tutorial, lilianweng2021diffusion}.

\subsection{Computational Considerations}

\subsubsection{GANs}
Despite their theoretical elegance, GANs are notoriously difficult to train in practice \citep{goodfellow2016nips, salimans2016improved}. One central issue is training instability: since the generator and discriminator are simultaneously optimized in a minimax game, updates that improve one network can destabilize the other. This often leads to oscillatory dynamics or divergence, rather than convergence to a Nash equilibrium—especially in high-dimensional, non-convex settings \citep{arjovsky2017towards}. Another common pathology is \emph{mode collapse}, where the generator produces limited and repetitive outputs, capturing only a subset of the data distribution. This typically arises when the generator overfits to fooling the current state of the discriminator, rather than learning a general mapping.

A further challenge lies in maintaining a balance between the generator and the discriminator. If the discriminator becomes too accurate, the generator receives vanishing gradients; if the generator dominates, it may exploit weaknesses in the discriminator instead of improving output fidelity. Addressing this requires careful tuning of architectures, learning rates, and optimization strategies. Practical remedies include architectural heuristics, such as those introduced in DCGAN \citep{radford2015unsupervised}, and regularization techniques like one-sided label smoothing. Evaluation is also nontrivial: GANs do not optimize a tractable likelihood, so metrics like the Inception Score (IS) and Fréchet Inception Distance (FID) are commonly used, though each has known limitations \citep{salimans2016improved}.

\subsubsection{DDPMs}
DDPMs are widely recognized for their stable training dynamics and their ability to generate diverse, high-quality samples. However, these benefits come at a significant computational cost. Training involves simulating millions of forward and reverse diffusion steps, and sampling is inherently sequential: each generated example requires \( T \) full passes through the denoising network. As a result, DDPMs are considerably slower at inference time compared to single-step generative models such as GANs \citep{nichol2021improved, song2020denoising, dhariwal2021diffusion}.

These trade-offs, namely training efficiency versus sample quality and diversity, highlight the need for alternative generative modeling frameworks that offer a better balance between computational demands, interpretability, and generative performance.

\subsubsection{GANs vs. DDPMs}
The differences between GANs and DDPMs are substantial, both in architecture and in training philosophy. GANs learn a direct mapping from latent noise to data through an adversarial game between a generator and a discriminator. This setup allows for fast sample generation, but it often suffers from training instability, mode collapse, and sensitivity to hyperparameter tuning \citep{goodfellow2014generative, salimans2016improved}.

In contrast, DDPMs learn to generate data via an iterative denoising process, guided by a stable, likelihood-inspired objective. While this tends to produce samples with greater diversity and realism, the generation process is much slower due to the required sequence of denoising steps \citep{nichol2021improved, ho2020denoising}.

Despite their conceptual differences, both frameworks rely heavily on deep neural networks as function approximators. In practice, the idealized objectives—such as the optimal discriminator in GANs or perfect noise prediction in DDPMs—are difficult to achieve. This mismatch between theoretical guarantees and empirical behavior contributes to the practical challenges each model faces \citep{salimans2016improved, ho2020denoising, nichol2021improved}.

\section{Support Points: Theory and Algorithm}

Support points offer a nonparametric approach to approximate a continuous probability distribution  \( F \) using a finite set of representative points. The resulting point set balances fidelity to the distribution with diversity among elements, making support points particularly attractive for tasks that require sample efficiency, interpretability, and structured coverage of the input space.

\subsection{Mathematical Foundations of Support Points}

The central idea behind support points is to construct a discrete probability measure that approximates a target distribution \( F \) as closely as possible under a rigorous statistical criterion. This criterion relies on the \emph{energy distance}, a metric between probability distributions, defined for two distributions \( F \) and \( G \) on a subset \( \mathcal{X} \subseteq \mathbb{R}^d \), both with finite means, as \citep{szekely2005new, szekely2013energy, mak2018support} 
\begin{equation}
E(F, G) = 2\, \mathbb{E} \| \mathbf{X} - \mathbf{Y} \|_2 - \mathbb{E} \| \mathbf{X} - \mathbf{X}' \|_2 - \mathbb{E} \| \mathbf{Y} - \mathbf{Y}' \|_2,
\label{eq:energy_distance_def}
\end{equation}
where \( \mathbf{X}, \mathbf{X}' \overset{\text{i.i.d.}}{\sim} G \), \( \mathbf{Y}, \mathbf{Y}' \overset{\text{i.i.d.}}{\sim} F \), and \( \|\cdot\|_2 \) denotes the Euclidean norm. This metric can be interpreted as a generalized notion of potential energy between distributions and was originally proposed for multivariate, nonparametric goodness-of-fit testing \citep{szekely2004testing}.

Importantly, \( E(F, G) \geq 0 \) for all \( F, G \), and \( E(F, G) = 0 \) if and only if \( F = G \). These properties make the energy distance a valid statistical metric and a natural criterion for distributional approximation.

\subsection{Definition of Support Points}

Given a target distribution \( F \) and a desired number of support points \( n \), the goal is to construct a set of vectors \( \boldsymbol{\xi}_1, \ldots, \boldsymbol{\xi}_n \subset \mathcal{X} \subseteq \mathbb{R}^d \) that approximates \( F \) as closely as possible. Let \( \mathbf{S} \in \mathbb{R}^{d \times n} \) denote the matrix whose columns are the support points. These points are defined as the solution to the minimization problem
\begin{equation}
\mathbf{S} = \underset{\mathbf{A} \in \mathbb{R}^{d \times n}}{\operatorname{argmin}}\, E(F, \widehat{F}_n),\label{mspp}
\end{equation}
where \( \widehat{F}_n \) denotes the empirical distribution associated with the columns \( \mathbf{x}_1, \dots, \mathbf{x}_n \) of a candidate matrix \( \mathbf{A} \in \mathbb{R}^{d \times n} \)
\[
\widehat{F}_n(\mathbf{x}) = \frac{1}{n} \sum_{i=1}^n \mathbb{I}\{ \mathbf{x}_i \leq \mathbf{x} \}.
\]

Using the definition of the energy distance between two distributions \( F \) and \( G \), and applying it to the case \( G = \widehat{F}_n \), we obtain
\begin{align}
E(F, \widehat{F}_n) & = 
\frac{2}{n} \sum_{i=1}^n \mathbb{E}_{\mathbf{Y} \sim F} \left[ \| \mathbf{x}_i - \mathbf{Y} \|_2 \right] \nonumber\\
& \quad - \frac{1}{n^2} \sum_{i=1}^n \sum_{j=1}^n \| \mathbf{x}_i - \mathbf{x}_j \|_2 \nonumber\\
& \quad - \mathbb{E}_{\mathbf{Y}, \mathbf{Y}'} \left[ \| \mathbf{Y} - \mathbf{Y}' \|_2 \right].
\label{eq:sp_energy_full}
\end{align}

This expression follows from the fact that \( \widehat{F}_n \) is a discrete probability measure that assigns equal mass \( 1/n \) to each point \( \mathbf{x}_i \). Expectations under \( \widehat{F}_n \) therefore take the form
\[
\mathbb{E}_{\mathbf{X} \sim \widehat{F}_n} [h(\mathbf{X})] = \int h(\mathbf{x})\, d\widehat{F}_n(\mathbf{x}) = \frac{1}{n} \sum_{i=1}^n h(\mathbf{x}_i).
\]


Let \( \mathcal{P}_n(\mathbb{R}^d) \) denote the space of probability measures supported on \( n \) points. From this perspective, support points correspond to an optimal discrete measure \( \widehat{F}_n \in \mathcal{P}_n(\mathbb{R}^d) \) that minimizes the energy distance to a target measure \( F \in \mathcal{P}(\mathbb{R}^d) \).

This variational formulation highlights the interpretation of support points as the best approximation to \( F \) within a finite-dimensional class of discrete measures, using the geometry induced by the energy distance.

\subsubsection{Optimization: The Convex-Concave Procedure (CCP)}

Direct minimization of the objective function~\eqref{eq:sp_energy_full} presents two main challenges: (i) the expectation term \( \mathbb{E}_{\mathbf{Y} \sim F} \| \mathbf{x}_i - \mathbf{Y} \|_2 \) is generally intractable in closed form, and (ii) the overall function is non-convex due to the interplay between attractive and repulsive components.

To address these difficulties, \citet{mak2018support} propose a two-step strategy. First, the distribution \( F(\mathbf{y}) \) is approximated by an empirical reference measure \( \widehat{F}_N(\mathbf{y}) \), supported on a sample \( \{ \mathbf{y}_m \}_{m=1}^N \subset \mathbb{R}^d \). Letting \( \mathbf{P} = [\mathbf{y}_1, \dots, \mathbf{y}_N] \), the objective becomes the (unnormalized) empirical energy distance between \( \widehat{F}_n \) and \( \widehat{F}_N \)
\begin{align}
MC(\mathbf{A}; \mathbf{P}) &= 
\underbrace{
\frac{2}{nN} \sum_{i=1}^n \sum_{m=1}^N \| \mathbf{x}_i - \mathbf{y}_m \|_2
}_{\text{attraction term}}\nonumber\\
&\quad -
\underbrace{
\frac{1}{n^2} \sum_{i=1}^n \sum_{j=1}^n \| \mathbf{x}_i - \mathbf{x}_j \|_2
}_{\text{repulsion term}},
\label{eq:sp_objective_mc}
\end{align}
where \( \mathbf{A} = [\mathbf{x}_1, \dots, \mathbf{x}_n] \in \mathbb{R}^{d \times n} \) denotes the matrix of candidate points.

In the second step, the optimization is carried out via the Convex-Concave Procedure (CCP), an iterative method for minimizing functions expressed as the difference of two convex terms. The structure of~\eqref{eq:sp_objective_mc} naturally fits this framework:
\[
MC(\mathbf{A}; \mathbf{P}) = f_{\text{att}}(\mathbf{A}; \mathbf{P}) - f_{\text{rep}}(\mathbf{A}),
\]
where \( f_{\text{att}} \) corresponds to the attraction term, encouraging each point \( \mathbf{x}_i \) to move toward high-density regions of \( \widehat{F}_N \), and \( f_{\text{rep}} \) encodes a pairwise repulsion that spreads the points across the domain. Both components are convex, but their difference yields a non-convex objective.

At each CCP iteration, the concave part \( f_{\text{rep}} \) is linearized around the current iterate, resulting in a convex surrogate problem. Solving this surrogate yields an updated configuration of points. Conceptually, the iteration corresponds to approximating the gradient of the full objective by freezing the non-convexity, and updating each \( \mathbf{x}_i \) through an explicit expression involving the data and the remaining support points. This process is repeated from a suitable initialization to produce a sequence \( \mathbf{A}^{(0)}, \mathbf{A}^{(1)}, \dots \) that converges to a local minimum.

This formulation effectively minimizes \( E(\widehat{F}_N, \widehat{F}_n) \) up to an additive constant, yielding a principled and computationally tractable surrogate for the original objective \( E(F, \widehat{F}_n) \).

We now present Algorithm \ref{alg:sp_ccp} which approximates the support points matrix \( \mathbf{S} \in \mathbb{R}^{d \times n} \) that minimizes the objective in equation~\eqref{eq:sp_objective_mc}.

\begin{algorithm}
\caption{\texttt{sp.ccp}: Support Points via Cyclic Convex Procedure}\label{alg:sp_ccp}
\begin{algorithmic}[1]
\State \textbf{Input:} Dataset \( \mathbf{P} \in \mathbb{R}^{d \times N} \), initial candidate matrix \( \mathbf{A}^{(0)} \in \mathbb{R}^{d \times n} \), and convergence tolerance \( \epsilon \).
\State Compute initial cost: \( s^0 = MC(\mathbf{A}^{(0)}; \mathbf{P}) \).
\State Set iteration counter \( l \gets 0 \).
\Repeat
    \For{$i = 1, \dots, n$ \textbf{(in parallel)}}
        \State Update candidate point:
        \[
        \mathbf{x}_i^{(l+1)} \gets M_i(\mathbf{A}^{(l)}; \mathbf{P}) \qquad \text{(see equation~\eqref{eq1})}
        \]
    \EndFor
    \State Set \( \mathbf{A}^{(l+1)} \gets [\mathbf{x}_1^{(l+1)}, \dots, \mathbf{x}_n^{(l+1)}] \).
    \State Compute updated cost: \( s^{l+1} = MC(\mathbf{A}^{(l+1)}; \mathbf{P}) \).
    \State Increment iteration: \( l \gets l + 1 \).
\Until{ \( |s^{l} - s^{l-1}| < \epsilon \) }
\State \textbf{Output:} Final support point matrix \( \mathbf{S} = \mathbf{A}^{(l)} \).
\end{algorithmic}
\end{algorithm}

The mapping \( M_i \), which updates the \( i \)-th column, is defined using a reference matrix \( \mathbf{A}' = [\mathbf{x}_1', \dots, \mathbf{x}_n'] \in \mathbb{R}^{d \times n} \), where each \( \mathbf{x}_j' \) is treated as fixed during the current iteration. The updated value of \( \mathbf{x}_i \) is given by
\begin{align}
\mathbf{x}_i &= M_i\left( \mathbf{A}', \mathbf{P} \right) \nonumber \\
&= \frac{1}{q(\mathbf{x}_i', \mathbf{P})}
\left( \sum_{m=1}^N \frac{\mathbf{y}_m}{\|\mathbf{x}_i' - \mathbf{y}_m\|_2} + \frac{N}{n} \sum_{\substack{j=1 \\ j \neq i}}^n \frac{\mathbf{x}_i' - \mathbf{x}_j'}{\|\mathbf{x}_i' - \mathbf{x}_j'\|_2}
\right), \label{eq1}
\end{align}
where the normalizing factor is:
\[
q(\mathbf{x}_i', \mathbf{P}) = \sum_{m=1}^N \frac{1}{\|\mathbf{x}_i' - \mathbf{y}_m\|_2}.
\]

In the original formulation by \citet{mak2018support}, the initial matrix \( \mathbf{A}^{(0)} \in \mathbb{R}^{d \times n} \) is constructed by randomly selecting \( n \) columns from the reference matrix \( \mathbf{P} \). This strategy is simple and ensures that the initial candidate points lie within the domain of the data. However, it may lead to numerical instabilities when duplicate or nearly identical vectors are chosen, particularly in high-dimensional settings.

To address this, our implementation allows for alternative initialization schemes. By default, \( \mathbf{A}^{(0)} \) can be generated by drawing independent samples from a uniform distribution over the bounding box of \( \mathbf{P} \). This method avoids exact duplications and introduces greater spatial diversity in the initial configuration. The initialization strategy is configurable, enabling users to experiment with different options and select the one that yields the best empirical performance for their task.

Unlike generative models such as GANs or diffusion-based methods, which produce samples by transforming latent noise through a trained neural network, support points offer a nonparametric alternative. They approximate the target distribution directly by selecting a finite set of representative points that minimize a statistical distance. This makes them particularly appealing in applications where interpretability, sample efficiency, and structured coverage are critical—especially in data-limited or resource-constrained environments.

\section{Weighted Support Points}

The standard minimization problem, expression~\eqref{mspp}, and its associated algorithm converge deterministically to a single set of support points. While this property is desirable in approximation tasks, where a stable and representative summary of the distribution is required, it limits the method’s applicability in generative settings where diversity is essential.

To address this limitation, we introduce a simple yet effective modification to the optimization objective, motivated by the idea of replacing the target distribution \( F \) with a random probability measure \( \widetilde{F} \). This leads to a randomized version of the energy distance objective, which in turn produces multiple, diverse configurations of support points. We refer to these as \emph{weighted support points}, as each realization of \( \widetilde{F} \) corresponds to a weighted empirical measure supported on the original dataset.

\subsection{Random Measures}

In the classical formulation, support points approximate a fixed probability measure \( F \in \mathcal{P}(\mathbb{R}^d) \) by minimizing the energy distance to a discrete uniform measure supported on \( n \) points. While this is appropriate for summarization tasks, it limits the generation of diverse outputs, an essential feature in generative modeling.

To introduce stochasticity, we replace the fixed distribution \( F \) with a random probability measure \( \widetilde{F} \), supported on a finite reference set \( \{ \mathbf{y}_m \}_{m=1}^N \subset \mathbb{R}^d \). We consider random measures of the form
\begin{equation}
    \widetilde{F}_N(\mathbf{y}) = \sum_{m=1}^N w_m \, \mathbb{I}\{ \mathbf{y}_m \leq \mathbf{y} \},\label{rF}
\end{equation}
where \( \mathbf{w} = (w_1, \dots, w_N) \) is a vector of nonnegative weights summing to one. This general formulation encompasses several classical constructions
\begin{itemize}
    \item The empirical distribution \( \widehat{F}_N \), where \( w_m = 1/N \) for all \( m \),
    \item The Bayesian bootstrap of \citet{rubin1981bayesian}, where \( \mathbf{w} \sim \mathrm{Dirichlet}(1, \dots, 1) \),
    \item Truncated Dirichlet processes \citep{ferguson1973}, where the weights \( \mathbf{w} \) are generated via the stick-breaking construction of \citet{sb1994}.
\end{itemize}

Substituting \( F \) with a realization of \( \widetilde{F}_N \) in the energy distance objective yields a randomized variant of the support points algorithm. Each draw of \( \mathbf{w} \) defines a different optimization landscape, producing distinct configurations of representative points. We refer to these as \emph{weighted support points}. This approach offers a principled way to introduce controlled randomness into the method, enabling the generation of diverse outputs while maintaining fidelity to the data-generating distribution.

\subsection{Definition of Weighted Support Points}\label{wsp}

We define the \emph{weighted support points} as the solution to the following stochastic optimization problem:
\begin{equation*}
\mathbf{S}(\mathbf{w}) = \underset{\mathbf{A} \in \mathbb{R}^{d \times n}}{\operatorname{argmin}}\, E(\widetilde{F}_N, \widehat{F}_n),
\end{equation*}
where \( \widetilde{F}_N \) is a discrete random probability measure supported on a reference dataset \( \mathbf{P} = \{ \mathbf{y}_m \}_{m=1}^N \subset \mathbb{R}^d \), with weights \( \mathbf{w} = (w_1, \dots, w_N) \) randomly drawn and held fixed during each optimization run.

Ignoring the constant term that does not depend on \( \mathbf{A} \), the problem reduces to minimizing the following weighted empirical energy distance
\begin{align}
MC(\mathbf{A}; \mathbf{P}, \mathbf{w}) &=
\frac{2}{n} \sum_{i=1}^n \sum_{m=1}^N w_m \| \mathbf{x}_i - \mathbf{y}_m \|\nonumber\\
&- \frac{1}{n^2} \sum_{i=1}^n \sum_{j=1}^n \| \mathbf{x}_i - \mathbf{x}_j \|. \label{eq:mc_weighted}
\end{align}

Solving this optimization problem for multiple independent realizations of \( \mathbf{w} \) yields a diverse collection of representative point sets, each adapted to a different randomized view of the underlying data distribution.

\subsubsection{Optimization}

To derive the update rule for weighted support points, we compute the gradient of the objective function in equation~\eqref{eq:mc_weighted} with respect to each point \( \mathbf{x}_i \), assuming the weights \( \mathbf{w} = (w_1, \dots, w_N) \) are fixed during the optimization.

The gradient of the attraction term with respect to \( \mathbf{x}_i \) is:
\[
\frac{\mathrm{d}}{\mathrm{d} \mathbf{x}_i} \left(
\frac{2}{n} \sum_{k=1}^n \sum_{m=1}^N w_m \| \mathbf{x}_k - \mathbf{y}_m \|
\right)
= \frac{2}{n} \sum_{m=1}^N w_m \frac{\mathbf{x}_i - \mathbf{y}_m}{\| \mathbf{x}_i - \mathbf{y}_m \|}.
\]
The gradient of the repulsion term with respect to \( \mathbf{x}_i \) is:
\[
\frac{\mathrm{d}}{\mathrm{d} \mathbf{x}_i} \left(
\frac{1}{n^2} \sum_{k=1}^n \sum_{j=1}^n \| \mathbf{x}_k - \mathbf{x}_j \|
\right)
= \frac{2}{n^2} \sum_{\substack{j = 1 \\ j \neq i}}^n \frac{\mathbf{x}_i - \mathbf{x}_j}{\| \mathbf{x}_i - \mathbf{x}_j \|}.
\]
Setting the total gradient to zero yields the first-order condition
\[
\frac{2}{n} \sum_{m=1}^N w_m \frac{\mathbf{x}_i - \mathbf{y}_m}{\| \mathbf{x}_i - \mathbf{y}_m \|}
=
\frac{2}{n^2} \sum_{\substack{j = 1 \\ j \neq i}}^n \frac{\mathbf{x}_i - \mathbf{x}_j}{\| \mathbf{x}_i - \mathbf{x}_j \|}.
\]
Rearranging this expression leads to an explicit update rule. Let \( \mathbf{x}_i' \) denote the current value of \( \mathbf{x}_i \). Then the next iterate is given by
\begin{equation}
\mathbf{x}_i =
\frac{1}{q_i} \left(
\sum_{m=1}^N \frac{w_m \mathbf{y}_m}{\| \mathbf{x}_i' - \mathbf{y}_m \|}
+ \frac{1}{n} \sum_{\substack{j = 1 \\ j \neq i}}^n \frac{\mathbf{x}_i' - \mathbf{x}_j'}{\| \mathbf{x}_i' - \mathbf{x}_j' \|}
\right),
\label{eq:Mi_weighted}
\end{equation}
with normalization constant
\[
q_i = \sum_{m=1}^N \frac{w_m}{\| \mathbf{x}_i' - \mathbf{y}_m \|}.
\]

This fixed-point iteration is repeated until convergence to obtain a locally optimal configuration of weighted support points for the current realization of the weights \( \mathbf{w} \).

The optimization follows the same iterative structure as the original \texttt{sp.ccp} algorithm (Algorithm~\ref{alg:sp_ccp}), with two key modifications:
(i) the objective function is replaced by its weighted form~\eqref{eq:mc_weighted}, and 
(ii) the update rule \( M_i \) is modified accordingly, as shown in equation~\eqref{eq:Mi_weighted}. To efficiently evaluate the objective and update steps, we exploit symmetries and reuse intermediate computations using inner products and matrix caching. These optimizations significantly reduce cost in large-scale settings. Full details are provided in Appendix~\ref{ee}.

\subsection{Promoting Diversity via Random Subsetting and Symmetric Dirichlet Weights}\label{diversity}

In generative settings, we seek support points that are not only representative of the full dataset but also diverse across independent runs. When relying on a random measure \( \widetilde{F}_N \) defined over a fixed reference set \( \{\mathbf{y}_m\}_{m=1}^N \), there is an inherent trade-off between
\begin{itemize}
    \item \textbf{Structural representativeness}: keeping the full dataset accessible so that all reference points can be informative;
    \item \textbf{Sample diversity}: inducing run-to-run variability by emphasizing different subsets of the data in each realization.
\end{itemize}
Although (\ref{rF}) is random, under common constructions (e.g., Bayesian bootstrap or finite Dirichlet-based schemes) it is centered at the empirical distribution \( \widehat{F}_N \).

Two benchmarks illustrate opposite limitations. The \emph{Bayesian bootstrap} assigns weights to \emph{all} \(N\) atoms with \(\mathbf{w}\sim\mathrm{Dir}(1,\ldots,1)\), so \(\mathbb{E}[w_i]=1/N\) and \(\mathrm{Var}(w_i)=\frac{N-1}{N^2(N+1)}=\mathcal{O}(N^{-2})\). For large \(N\), the induced perturbation of empirical averages is small in absolute terms, and consecutive runs remain close to \(\widehat F_N\), limiting diversity. By contrast, a \emph{Dirichlet–process} draw supported on \(\{\mathbf{y}_1,\ldots,\mathbf{y}_N\}\) yields \(\mathbf{w}\sim\mathrm{Dir}(\alpha/N,\ldots,\alpha/N)\), with relative dispersion \(\mathrm{CV}\approx\sqrt{N/(\alpha+1)}\); for fixed \(\alpha\) this produces spiky allocations where a few atoms dominate, harming coverage and stability.

To strike a balance, we adopt a finite random measure built by random subsetting and symmetric–Dirichlet weighting calibrated by a target coefficient of variation (CV). Let \(I=\{i_1,\ldots,i_{N_0}\}\subset\{1,\ldots,N\}\) be a random subset sampled uniformly without replacement (the rule for choosing \(N_0\) is given below). Given \(I\), draw \(\mathbf{w}=(w_1,\ldots,w_{N_0})\sim\mathrm{Dirichlet}(\alpha,\ldots,\alpha)\) with total concentration \(\kappa=N_0\alpha\) calibrated via the target CV (see expression (\ref{CV})). The resulting random probability measure is
\[
\widetilde{F}_{N_0}(\mathbf{y})
\;=\;
\sum_{j=1}^{N_0} w_j\,\mathbb{I}\{\mathbf{y}_{i_j}\le \mathbf{y}\}.
\]
Because \(\mathbb{E}[w_j\mid I]=1/N_0\) for a symmetric Dirichlet and each index has equal inclusion probability under uniform subsetting, it follows that
\begin{equation}
\mathbb{E}_{I,\mathbf{w}}\big[\widetilde{F}_{N_0}(\mathbf{y})\big]
\;=\;
\widehat{F}_N(\mathbf{y}),
\label{eq:centering_property}
\end{equation}
as shown in Appendix~\ref{cm}. Hence \(\widetilde{F}_{N_0}\) is centered at \(\widehat{F}_N\), while its dispersion is controlled by \(\kappa\) (equivalently, by the chosen CV), which enhances diversity across runs without concentrating mass on a few atoms. The subsections below detail the subsetting rule and the CV–based calibration of \(\kappa\).

\subsubsection{Subset size and selection}
Our aim is to balance coverage of the reference set with between-run variability in the active atoms. We randomize the retained size while enforcing a minimum retention rate. Specifically, draw \(\theta\sim\mathrm{Unif}(0.7,0.9)\) and set
\[
N_0 \;=\; \max\bigl\{\lceil 0.6\,N\rceil,\ S\bigr\}, 
\qquad S \sim \mathrm{Binomial}(N,\theta).
\]
The binomial mechanism corresponds to independent thinning with inclusion probability \(\theta\). Conditional on \(\theta\), \(\mathbb{E}[S\mid\theta]=\theta N\) and \(\mathrm{Var}(S\mid\theta)=N\theta(1-\theta)\). For large \(N\) and \(\theta\in[0.7,0.9]\), the lower bound \(\lceil 0.6N\rceil\) is rarely active; it simply prevents overly aggressive pruning so the retained dictionary remains well anchored to the data geometry. Given \(N_0\), we sample \(I\) uniformly without replacement and work with \(\{\mathbf{y}_{i_1},\ldots,\mathbf{y}_{i_{N_0}}\}\). Randomizing \(\theta\) across runs induces additional variability in \(N_0\), which helps decorrelate successive realizations and mitigates near-identical configurations upon reinitialization.

\subsubsection{Symmetric Dirichlet weights and CV calibration}
On the selected subset, draw \(\mathbf{w}\sim\mathrm{Dirichlet}(\alpha,\ldots,\alpha)\) with \(\kappa=N_0\alpha>0\). Equivalently, sample \(h_j\sim\mathrm{Gamma}(\alpha,1)\) and set
\[
w_j \;=\; \frac{h_j}{\sum_{\ell=1}^{N_0} h_\ell},\qquad j=1,\ldots,N_0.
\]
For any \(\kappa>0\),
\[
\mathbb{E}[w_j]=\frac{1}{N_0},\qquad 
\mathrm{Var}(w_j)=\frac{N_0-1}{N_0^2(\kappa+1)}.
\]
Hence the dispersion relative to the mean \(1/N_0\) is
\begin{equation}
\mathrm{CV}\;:=\;\frac{\sqrt{\mathrm{Var}(w_j)}}{1/N_0}
\;=\;\sqrt{\frac{N_0-1}{\kappa+1}}.
\label{CV}
\end{equation}
Given a target \(\mathrm{CV}>0\), we invert to obtain
\[
\kappa \;=\; \frac{N_0-1}{\mathrm{CV}^2}-1,
\]
which is feasible whenever \(\mathrm{CV}<\sqrt{N_0-1}\). Larger \(\kappa\) (equivalently, larger \(\alpha\)) yields more even weights; smaller \(\kappa\) yields more irregular allocations while preserving exchangeability within the subset.

The complete strategy is described in Appendix \ref{cm}.

\section{Examples}

\subsection{Example: MNIST Digits 6 and 9}

To illustrate the proposed method, we applied the weighted support point algorithm to a filtered subset of the MNIST dataset containing only images of the digits ``6'' and ``9''. Each grayscale image was flattened into a column vector of dimension \(784 = 28 \times 28\), resulting in a reference matrix \( \mathbf{P} \in \mathbb{R}^{784 \times 11867} \).

\begin{figure}[H]
  \centering
    \centering
    \includegraphics[width=0.9\linewidth]{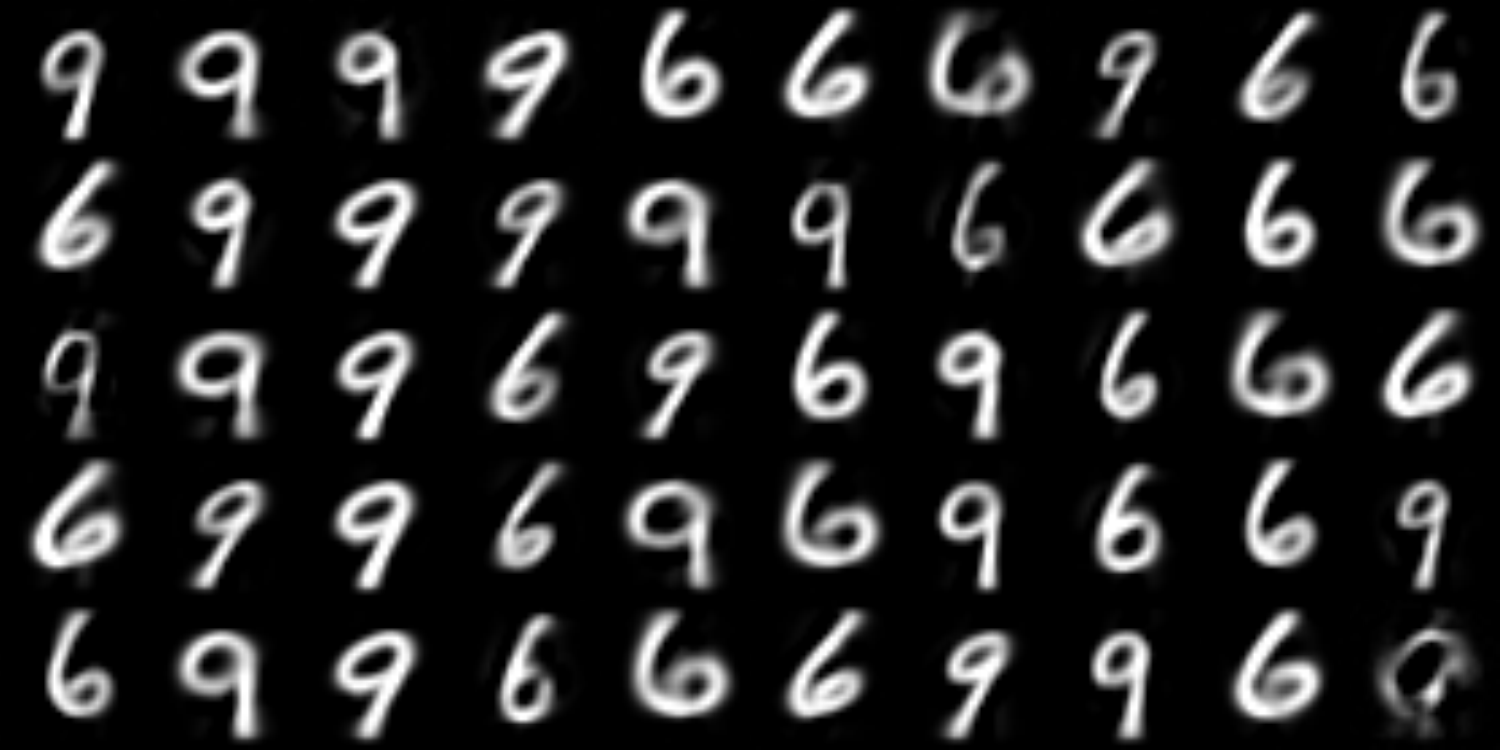}
    \caption{Five independent runs of the algorithm, each row is an independent run of size $n = 10$.}
    \label{fig:support-mnist-9}
  
  \vspace{0.5cm}

    \centering
    \includegraphics[width=0.9\linewidth]{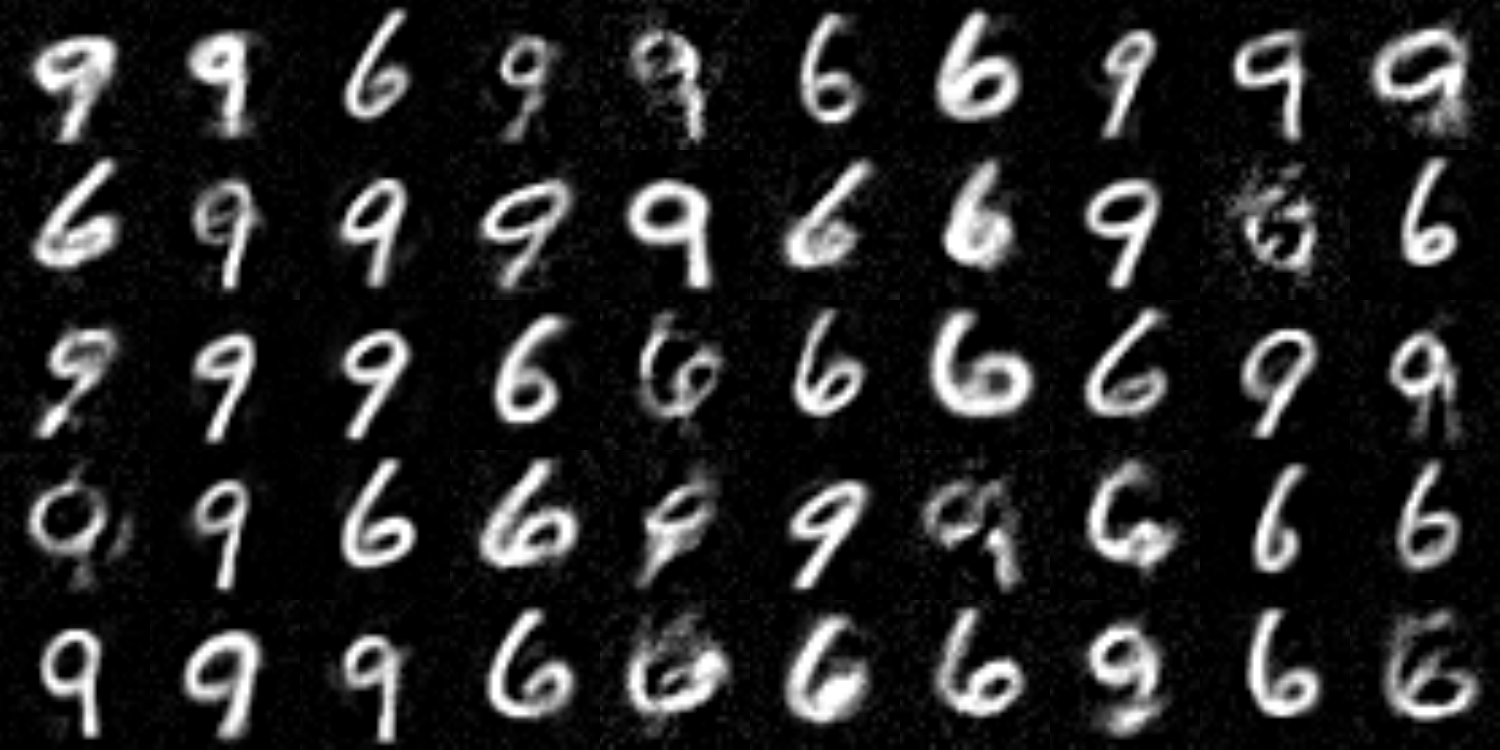}
    \caption{A single run of the algorithm ($n = 50$).}
    \label{fig:support-mnist-950}
\end{figure}

We generated \( n = 10 \) weighted support points in each of five independent runs (Figure~\ref{fig:support-mnist-9}), and also generated \( n = 50 \) weighted support points in a single run (Figure~\ref{fig:support-mnist-950}). In each case, the procedure was as follows:
\begin{itemize}
  \item The initial candidate matrix \( \mathbf{A}^{(0)} \in \mathbb{R}^{784 \times n} \) was sampled independently from a uniform distribution over \([0,1]\).
  \item A sparse random measure was generated using Algorithm\\~\texttt{gen.rmeasure} (Appendix~\ref{ee2}) with \( \mbox{CV} = 0.4 \), yielding a truncated set of atoms \( \widetilde{\mathbf{P}} \subset \mathbf{P} \) and normalized weights \( \widetilde{\mathbf{w}} \).
  \item The weighted support point algorithm \texttt{esp.ccp-w} (Appendix~\ref{ee}) was run with a maximum of 1,000 iterations and a convergence tolerance of \(10^{-5}\). Column-wise updates were parallelized across 16 CPU cores. Each run with \(n = 10\) completed in approximately 1 minute, while the run with \(n = 50\) took around 4 minutes.
  \item Final pixel values were clipped to \([0, 255]\) and rounded to the nearest integer.
\end{itemize}

Each row of Figure~\ref{fig:support-mnist-9} illustrates the diversity across independent runs, while Figure~\ref{fig:support-mnist-950} shows no discernible pattern, indicating that different representative images are consistently selected. This demonstrates that the stochasticity introduced by the random measure effectively promotes variability while preserving structure and legibility.

\subsubsection{Empirical comparisons}
\begin{figure}[H]
  \centering
    \centering
    \includegraphics[width=0.8\linewidth]{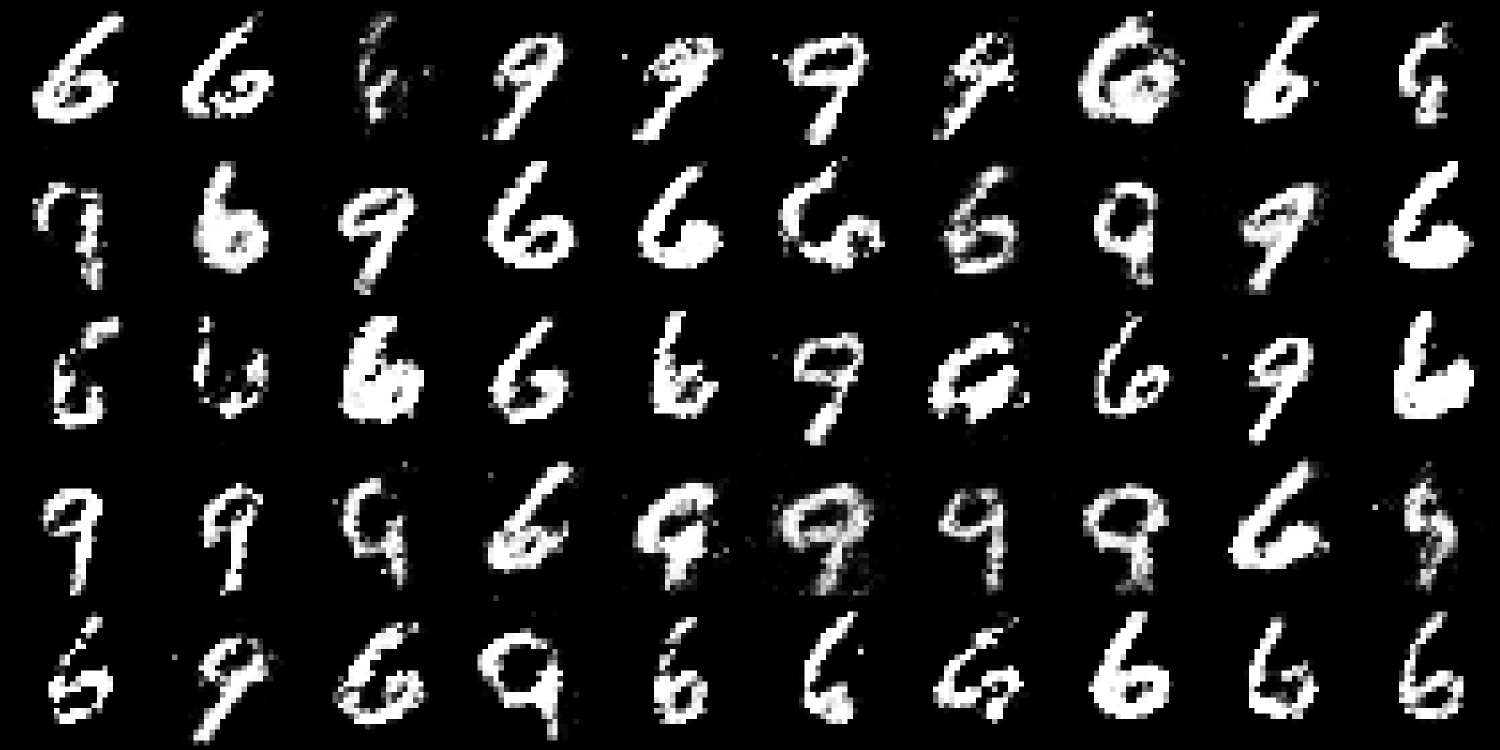}
    \caption{50 samples generated by the GANs.}
    \label{fig:gan_samples}
  
  \vspace{0.5cm}

    \centering
    \includegraphics[width=0.8\linewidth]{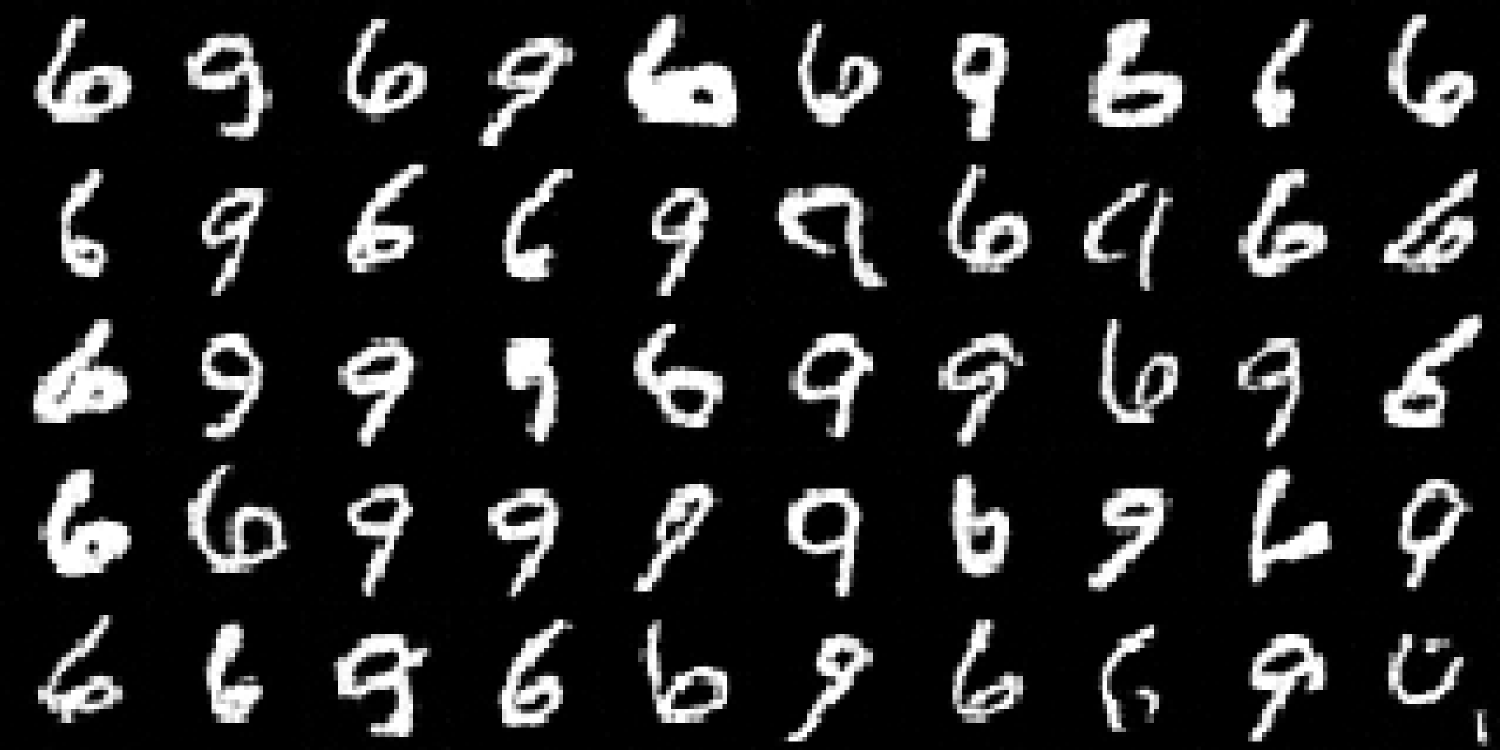}
    \caption{50 samples generated by the DDPMs.}
    \label{fig:ddpm_samples}
\end{figure}

To enable an empirical comparison, we generated samples of digits ``6'' and ``9'' using two neural-network-based generative approaches: a GANs (Figure~\ref{fig:gan_samples}) and a DDPMs (Figure~\ref{fig:ddpm_samples}). 
Both models were trained on the same dataset used for the weighted support points strategy.  

For the GANs, we used a lightweight fully connected architecture with spectral normalization in the discriminator. The generator maps a 100-dimensional noise vector to a $28\times 28$ grayscale image through three linear layers with ReLU activations and a final \texttt{Tanh} output. The discriminator processes flattened $28\times 28$ images through two linear layers with LeakyReLU activations and a final linear output for real/fake logits. Training used the Adam optimizer ($\mathrm{lr}=2\times 10^{-4}$, $\beta_1=0.5$, $\beta_2=0.999$), binary cross-entropy loss, batch size 128, label smoothing for real labels (0.9), $k=2$ discriminator updates per generator update, and up to 100 epochs. Once trained, the model generates images in milliseconds, producing recognizable digits 6 and 9 with relatively clean shapes.

For the DDPMs, we used a lightweight U-Net adapted to $32 \times 32$ grayscale images as the noise predictor at each diffusion step. 
Training employed the Adam optimizer ($\mathrm{lr}=2\times 10^{-5}$), batch size 64, and 5,000 epochs, minimizing the mean squared error (MSE) between true and predicted noise under a linear noise schedule with $T = 1,000$ steps. Compared to the GANs, training and image generation are considerably slower, but the resulting samples show smoother digit shapes and fewer visual artifacts, demonstrating the advantages of the diffusion-based approach even with a simple architecture.

Next, we compare these results with those obtained using our proposed weighted support points approach. As shown in Figures~\ref{fig:support-mnist-9} and \ref{fig:support-mnist-950}, the method selects clean and aesthetically pleasing representative digits ``6'' and ``9''. Although slightly blurred due to the use of distance-based optimization rather than pixel-level generative modeling, the outputs remain legible and structurally coherent. In contrast, the GAN (Figure~\ref{fig:gan_samples}) produces recognizable digits almost immediately after training, but with some speckled artifacts. The DDPMs (Figure~\ref{fig:ddpm_samples}) yields smoother results with fewer artifacts, though it occasionally generates distorted shapes, still clearly identifiable as ``6'' or ``9''. Overall, weighted support points offer a principled, interpretable, and computationally efficient alternative to neural generative models, producing diverse and stable samples without the need for heavy training.

A key methodological difference lies in how the data are represented and processed. Neural-network-based models (GANs and DDPMs) operate on images as multidimensional arrays, or \emph{tensors}, which encode batches of images along with their channels and spatial dimensions. These models also require dataset-specific transformations tightly coupled to their architectures and training procedures. In contrast, the weighted support points approach works directly on vectorized image representations, avoiding architecture-dependent preprocessing and enabling distance-based criteria to be applied in their native form.

It is worth noting that both the GANs and DDPMs can, in principle, achieve arbitrarily high performance by increasing the complexity of their architectures and training procedures. Such enhancements typically yield sharper and more realistic digits but at a substantial computational cost. In contrast, our weighted support points approach deliberately avoids neural architectures, focusing instead on interpretable and efficient sample generation.

\subsection{Example: CelebA-HQ Faces}

As a second example, we apply our algorithm to a subset of the CelebA-HQ dataset, which contains 27,000 high-quality color images of celebrity faces captured under unconstrained conditions. 
Each image has a resolution of $256 \times 256$ pixels and three color channels (RGB), yielding $256 \times 256 \times 3 = 196{,}608$ pixel values per image. 
Flattening all images into column vectors and stacking them produces a matrix
\[
\mathbf{P} \in \mathbb{R}^{196{,}608 \times 27{,}000},
\]
with over 5.3 billion entries. 
Storing this matrix in double precision (8 bytes per entry) would require
\[
196{,}608 \times 27{,}000 \times 8 \approx 39.3~\text{GB},
\]
which is infeasible on standard hardware.

\begin{figure}[H]
  \centering
\includegraphics[width=0.8\textwidth]{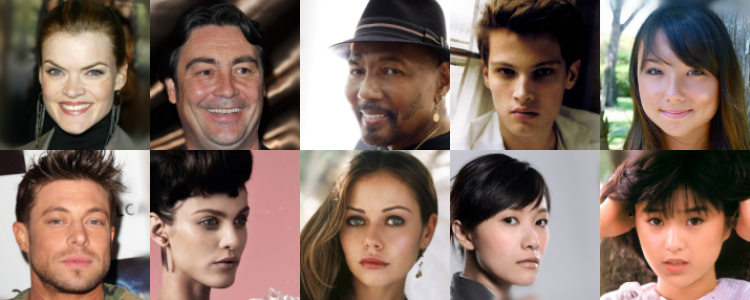}
  \caption{Sample of 10 downscaled CelebA-HQ face images selected from a subset of 10,000.}
  \label{fig:celebahq-original}
\end{figure}

To reduce memory requirements, we downscale each image to $144 \times 144$ pixels while preserving the RGB channels. 
Each image then becomes a vector in $\mathbb{R}^{62{,}208}$, and a random subset of 10,000 images is selected to form the reduced matrix $\mathbf{P} \in \mathbb{R}^{62{,}208 \times 10{,}000}$. 
A sample of ten resized images is shown in Figure~\ref{fig:celebahq-original}.

\begin{figure}[H]
  \centering
  \includegraphics[width=0.8\textwidth]{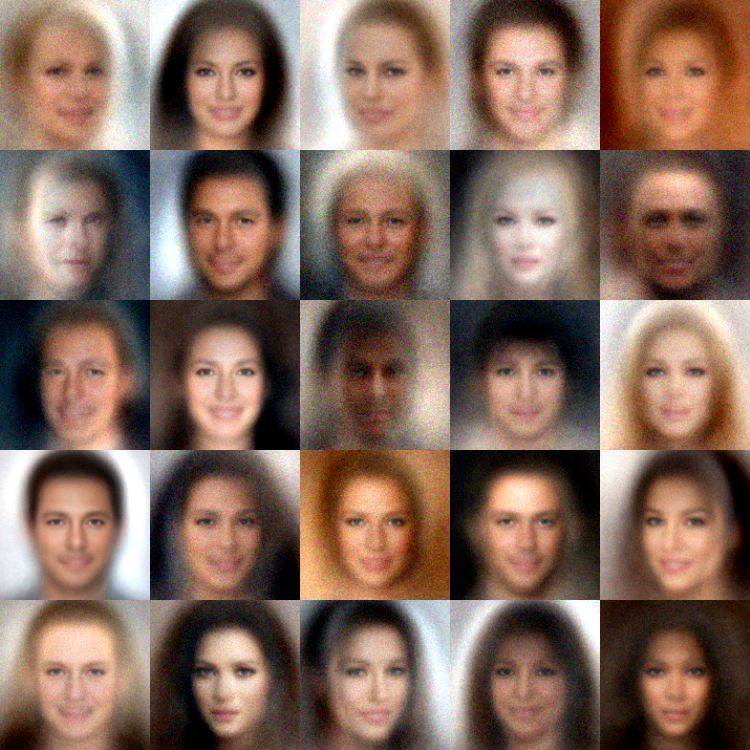}
  \caption{25 Weighted support points computed from a reduced subset of the CelebA-HQ dataset.}
  \label{fig:celebahq-support}
\end{figure}

We now apply the same methodology as for the MNIST Digits ``6'' and ``9'' to the reduced CelebA-HQ dataset. In a single run, we computed 25 weighted support points using a random measure \( \widetilde{F}_{N_0} \) constructed setting $\mbox{CV} = 0.4$ and via algorithm \texttt{gen.rmeasure} in Appendix \ref{ee2}. The optimization algorithm \texttt{esp.ccp-w} was initialized with a uniformly sampled matrix. The execution required approximately 3 hours to converge under tolerance \( \epsilon = 10^{-4} \) and a maximum of 1,000 iterations. Figure~\ref{fig:celebahq-support} displays the resulting weighted support points. 

Each row corresponds to a different run, and while some of the generated faces appear visually similar across runs, closer inspection reveals subtle but meaningful differences, such as variations in facial orientation or lighting. This suggests that the stochastic mechanism introduced by the random measure promotes diversity without compromising structural consistency. The observed variability confirms the method’s ability to explore different representative configurations while remaining faithful to the overall data distribution.

\section{Conclusion}

We introduced a generative modeling framework based on \emph{weighted support points} driven by a finite random measure built from \emph{random subsetting} and \emph{symmetric–Dirichlet} weights calibrated by a target coefficient of variation (CV) (see Section~\ref{diversity}). This construction yields two complementary sources of stochasticity, subset selection and weight randomness, while remaining \emph{centered} at the empirical distribution (expression (\ref{eq:centering_property}) and Appendix \ref{cm}). Calibrating the total concentration via the CV identity (\ref{CV}) provides explicit control over dispersion, preventing mass from collapsing onto a few atoms and promoting genuine run–to–run diversity.

Empirically, the method produces high–quality and diverse representatives on MNIST and CelebA–HQ at a fraction of the computational cost of neural approaches. The optimization is geometry–aware, the random measure is simple and transparent, and the overall pipeline is memory–efficient because it operates on a subset of atoms of size \(N_0\). In contrast to standard centering schemes, such as the Bayesian bootstrap, which perturbs all \(N\) atoms only mildly in absolute terms for large \(N\), or Dirichlet–process draws on a fixed support, which often induce overly spiky allocations for fixed \(\alpha\) our CV-calibrated subsetting strategy strikes a principled balance between structure and diversity.

Beyond the present results, the framework invites several extensions: stratified or geometry–aware subsetting (for example, \(k\)–center or determinantal selection), data–driven calibration of the CV, and finite–sample guarantees linking CV and diversity of the resulting support–point configurations. Overall, weighted support points with CV–calibrated symmetric–Dirichlet random measures constitute a practical, interpretable, and scalable alternative for generative modeling when stability, transparency, or limited computational resources are paramount.

\section*{Supplementary Material and Data Availability}

All code and datasets used to implement the algorithms and generate the experimental results presented in this paper is available as supplementary material. 
This includes the R implementation of the weighted support points algorithm, the sampling routine for truncated Dirichlet processes, and scripts for replicating the MNIST and CelebA-HQ experiments. 
Additionally, Python scripts for training and generating samples with both a simple GANs and a DDPMs on MNIST digits “6” and “9” are provided 
to facilitate reproducibility of the neural network–based comparison.

\section*{Funding}
The authors did not receive any specific funding for this work.

\section*{Conflict of Interest}
The authors declare that there is no conflict of interest.

\bibliographystyle{apalike}  
\bibliography{references}

\appendix
\section{Appendix A: Efficient Evaluation for Weighted Support Points}\label{ee}

The two main computational challenges in our method are evaluating the weighted cost function \( MC(\mathbf{A}^{(l)}; \mathbf{P}, \mathbf{w}) \) and updating the support points. For high-dimensional datasets, such as when \( \mathbf{P} \) contains images, naive implementations become computationally expensive. This section presents an efficient strategy that leverages algebraic identities, matrix operations, and symmetry to reuse intermediate computations and accelerate both cost evaluation and optimization updates.

In the repulsion term, we exploit symmetry and the identity \( \| \mathbf{x}_i^{(l)} - \mathbf{x}_j^{(l)} \|_2 = \| \mathbf{x}_j^{(l)} - \mathbf{x}_i^{(l)} \|_2 \). This allows us to compute each distance only once and scale appropriately. The weighted cost function, where the weights are placed on the reference atoms \( \mathbf{y}_m \), takes the form
\begin{align}
MC(\mathbf{A}^{(l)}; \mathbf{P}, \mathbf{w}) &=
\frac{2}{n} \sum_{i=1}^{n} \sum_{m=1}^{N} w_m \left\| \mathbf{x}_i^{(l)} - \mathbf{y}_m \right\|_2 \nonumber\\
&\quad - \frac{2}{n^2} \sum_{i=1}^{n} \sum_{j=1}^{i-1} \left\| \mathbf{x}_i^{(l)} - \mathbf{x}_j^{(l)} \right\|_2.
\label{mc-weighted-symmetric}
\end{align}

To compute the pairwise distances efficiently, we rely on algebraic expansions of squared Euclidean norms. These allow us to express distances in terms of squared norms and inner products:
\begin{align*}
\| \mathbf{x}_i^{(l)} - \mathbf{y}_m \|_2^2 &= \|\mathbf{x}_i^{(l)}\|_2^2 + \|\mathbf{y}_m\|_2^2 - 2\, {\mathbf{x}_i^{(l)}}^\top \mathbf{y}_m,\\
\| \mathbf{x}_i^{(l)} - \mathbf{x}_j^{(l)} \|_2^2 &= \|\mathbf{x}_i^{(l)}\|_2^2 + \|\mathbf{x}_j^{(l)}\|_2^2 - 2\, {\mathbf{x}_i^{(l)}}^\top \mathbf{x}_j^{(l)}.
\end{align*}

These expressions can be evaluated efficiently using matrix operations. To structure the computation, we distinguish between static and dynamic quantities:
\textit{static quantities} are computed once and reused across all iterations, while \textit{dynamic quantities} depend on the current value of \( \mathbf{A}^{(l)} \) and are updated in each iteration.

The static component consists of the squared norms of the data points:
\[
\mathbf{r}_p = \left( \|\mathbf{y}_1\|_2^2, \dots, \|\mathbf{y}_N\|_2^2 \right)^\top \in \mathbb{R}^N.
\]

At iteration \( l \), the dynamic cache \(\mathcal{C}^{(l)} = \texttt{Cache}(\mathbf{A}^{(l)}, \mathbf{P}, \mathbf{r}_p)\) includes:
\[
\mathcal{C}^{(l)} = \left\{
\begin{aligned}
& D_{xp}^{(l)} \in \mathbb{R}^{n \times N}, \quad
  D_{xp}^{(l)}[i,m] = \sqrt{ r_{x,i}^{(l)} + r_{p,m} - 2\, G_{xp}^{(l)}[i,m] }, \\
& D_{xx}^{(l)} \in \mathbb{R}^{n \times n}, \quad
  D_{xx}^{(l)}[i,j] = \sqrt{ r_{x,i}^{(l)} + r_{x,j}^{(l)} - 2\, G_{xx}^{(l)}[i,j] }, \\
& \mathrm{S}_{xp}^{(l)}(\mathbf{w}) = \sum_{i=1}^n \sum_{m=1}^N w_m \, D_{xp}^{(l)}[i,m] \in \mathbb{R}, \\
& \mathrm{S}_{xx}^{(l)} = \sum_{i=1}^{n} \sum_{j=1}^{i-1} \left\| \mathbf{x}_i^{(l)} - \mathbf{x}_j^{(l)} \right\|_2 \in \mathbb{R}
\end{aligned}
\right.
\]
where \( r_{x,i}^{(l)} = \| \mathbf{x}_i^{(l)} \|_2^2 \), and the inner product matrices are
\[
G_{xp}^{(l)} = \mathbf{A}^{(l)\top} \mathbf{P} \in \mathbb{R}^{n \times N}, \qquad
G_{xx}^{(l)} = \mathbf{A}^{(l)\top} \mathbf{A}^{(l)} \in \mathbb{R}^{n \times n}.
\]

To improve numerical stability in the evaluation of \(D_{xp}^{(l)}\) and \(D_{xx}^{(l)}\),
we replace the standard Euclidean norm with a smooth regularized form
\[
\sqrt{\max(d^2, 0) + \varepsilon^2},
\]
where \(\varepsilon\) is scaled to the problem as 
\(\varepsilon = c \sqrt{\texttt{.Machine\$double.eps}} \cdot s\).
Here, \(s\) denotes a characteristic norm of the current configuration and \(c\) is a small constant.
This scale-aware regularization mitigates instabilities when points coincide or are extremely close,
ensures gradients remain well-behaved, and avoids the inefficiencies that can arise from overly conservative numerical adjustments, while preserving the efficiency of the underlying vectorized matrix-based computation.

These cached quantities are used both to evaluate the cost function and to compute the update steps efficiently. The final form of the cost function becomes
\begin{equation*}
MC(\mathcal{C}^{(l)}; \mathbf{w}) =
\frac{2}{n} \, \mathrm{S}_{xp}^{(l)}(\mathbf{w}) \ -\ \frac{2}{n^2} \, \mathrm{S}_{xx}^{(l)}.
\end{equation*}

The optimization step is also accelerated using the cached matrices. Each candidate point is updated as:
\begin{align*}
\mathbf{x}_i^{(l+1)} &= M_i(i;\, \mathbf{A}^{(l)}, \mathbf{P}, \mathcal{C}^{(l)}, \mathbf{w}) \\
 &= \frac{1}{q_i^{(l)}} \left(
\sum_{m=1}^{N} 
\frac{w_m \mathbf{y}_m}{D_{xp}^{(l)}[i,m]}
+ \frac{1}{n} \sum_{\substack{j=1 \\ j \neq i}}^n 
\frac{\mathbf{x}_i^{(l)} - \mathbf{x}_j^{(l)}}{D_{xx}^{(l)}[i,j]}\right),
\end{align*}
where the normalization constant is
\[
q_i^{(l)} = \sum_{m=1}^{N} \frac{w_m}{D_{xp}^{(l)}[i,m]}.
\]

The key matrix operations, such as computing \( G_{xp}^{(l)} = \mathbf{A}^{(l)\top} \mathbf{P} \) and \( G_{xx}^{(l)} = \mathbf{A}^{(l)\top} \mathbf{A}^{(l)} \), are highly optimized in standard linear algebra libraries (e.g., BLAS level-3).

\begin{algorithm}
\caption{\texttt{esp.ccp-w}: Efficient Weighted Support Points via Cyclic Convex Procedure}
\begin{algorithmic}[1]
\State \textbf{Input:} Dataset \( \mathbf{P} \in \mathbb{R}^{d \times N} \); initial configuration \( \mathbf{A}^{(0)} \in \mathbb{R}^{d \times n} \); weights \( \mathbf{w} = (w_1, \dots, w_N)  \); convergence tolerance \( \epsilon \).
\State Precompute static norms: \( \mathbf{r}_p = \left( \|\mathbf{y}_1\|^2, \dots, \|\mathbf{y}_N\|^2 \right)^\top \in \mathbb{R}^N \).
\State Initialize cache: \( \mathcal{C}^{(0)} \gets \texttt{Cache}(\mathbf{A}^{(0)}, \mathbf{P}, \mathbf{r}_p) \).
\State Evaluate initial cost: \( s^0 \gets MC(\mathcal{C}^{(0)}, \mathbf{w}) \).
\State Set iteration counter: \( l \gets 0 \).
\Repeat
    \For{\( i = 1 \) to \( n \) \textbf{(in parallel)}}
        \State Update point: 
        \[
        \mathbf{x}_i^{(l+1)} \gets M_i(i;\, \mathbf{A}^{(l)}, \mathbf{P}, \mathcal{C}^{(l)}, \mathbf{w})
        \]
    \EndFor
    \State Aggregate updates: \( \mathbf{A}^{(l+1)} \gets [\mathbf{x}_1^{(l+1)}, \dots, \mathbf{x}_n^{(l+1)}] \).
    \State Update cache: \( \mathcal{C}^{(l+1)} \gets \texttt{Cache}(\mathbf{A}^{(l+1)}, \mathbf{P}, \mathbf{r}_p) \).
    \State Re-evaluate cost: \( s^{l+1} \gets MC(\mathcal{C}^{(l+1)}, \mathbf{w}) \).
    \State Increment: \( l \gets l + 1 \).
\Until{\( |s^l - s^{l-1}| < \epsilon \)}
\State \textbf{Output:} Weighted support point matrix \( \mathbf{Sw} = \mathbf{A}^{(l)} \).
\end{algorithmic}
\end{algorithm}

\section{Appendix B: Sampling a Random Measure via Random Subsetting and Symmetric Dirichlet Weights}\label{ee2}

This appendix (see Algorithm \ref{rmeasure} below) describes the random subsetting and symmetric Dirichlet approach described in Section~\ref{diversity}. The procedure draws a subset of atoms to introduce structured sparsity and then assigns Dirichlet–symmetric weights calibrated by a target coefficient of variation (CV). The resulting pair \((\widetilde{\mathbf{P}}, \widetilde{\mathbf{w}})\) defines a discrete random measure that is centered at the empirical distribution while allowing explicit control of weight dispersion and improved run–to–run diversity.

\begin{algorithm}
\caption{\texttt{gen.rmeasure}: Random Subsetting + Symmetric–Dirichlet Weights (CV–calibrated)}\label{rmeasure}
\begin{algorithmic}[1]
\State Input: reference set \(\mathbf{P}=\{\mathbf{y}_1,\ldots,\mathbf{y}_N\}\subset\mathbb{R}^d\); target coefficient of variation \(\mathrm{CV}>0\); subsetting rule parameters
\State Draw \(\theta \sim \mathrm{Unif}(0.7,0.9)\) and set the retained size
\[
N_0 \leftarrow \max\{\lceil 0.6 N\rceil,\ S\},\qquad S\sim\mathrm{Binomial}(N,\theta)
\]
(If a fixed subset size is preferred, set \(N_0\) directly.)
\State Sample a subset of indices \(I=\{i_1,\ldots,i_{N_0}\}\subset\{1,\ldots,N\}\) uniformly without replacement and form the reduced dictionary \(\widetilde{\mathbf{P}}=\{\mathbf{y}_{i_1},\ldots,\mathbf{y}_{i_{N_0}}\}\)
\State Calibrate the total concentration \(\kappa\) from the target CV via
\[
\kappa \leftarrow \frac{N_0-1}{\mathrm{CV}^2}-1
\]
and set the per–component Dirichlet parameter \(\alpha \leftarrow \kappa/N_0\) \quad (feasible when \(\mathrm{CV}<\sqrt{N_0-1}\))
\State Sample weights on the subset: draw \(h_j\sim\mathrm{Gamma}(\alpha,1)\) for \(j=1,\ldots,N_0\) and set
\[
\widetilde{w}_j \leftarrow \frac{h_j}{\sum_{\ell=1}^{N_0} h_\ell},\qquad j=1,\ldots,N_0
\]
\State Output: atom set \(\widetilde{\mathbf{P}}\) and weights \(\widetilde{\mathbf{w}}=(\widetilde{w}_1,\ldots,\widetilde{w}_{N_0})\)
\end{algorithmic}
\end{algorithm}

\section{Appendix C: Centered 
Random Measure}\label{cm}

Define the random finite measure
\[
\widetilde{F}_{N_0}(A)\;=\;\sum_{j=1}^{N_0} w_j\,\delta_{\mathbf{y}_{i_j}}(A),
\qquad A\subset\mathbb{R}^d,
\]
where \(I=\{i_1,\ldots,i_{N_0}\}\) is sampled uniformly without replacement from \(\{1,\ldots,N\}\), and \(\mathbf{w}=(w_1,\ldots,w_{N_0})\sim\mathrm{Dirichlet}(\alpha,\ldots,\alpha)\) (possibly with \(\alpha\) depending on \(N_0\) through the CV calibration). Then
\[
\mathbb{E}\big[\widetilde{F}_{N_0}(A)\big]
\;=\;
\mathbb{E}_{N_0}\Big\{
\mathbb{E}_{I\,|\,N_0}\big[
\mathbb{E}_{\mathbf{w}\,|\,I,N_0}\big[\widetilde{F}_{N_0}(A)\big]
\big]
\Big\}.
\]
Since the Dirichlet is symmetric, \(\mathbb{E}[w_j\,|\,I,N_0]=1/N_0\) for all \(j\), hence
\[
\mathbb{E}_{\mathbf{w}\,|\,I,N_0}\big[\widetilde{F}_{N_0}(A)\big]
=
\frac{1}{N_0}\sum_{j=1}^{N_0}\delta_{\mathbf{y}_{i_j}}(A)
=
\frac{1}{N_0}\sum_{m=1}^{N}\mathbf{1}\{m\in I\}\,\delta_{\mathbf{y}_m}(A).
\]
Conditionally on \(N_0\), uniform sampling without replacement gives
\(\mathbb{P}(m\in I\,|\,N_0)=N_0/N\), so
\begin{align*}
\mathbb{E}_{I\,|\,N_0}\!\left[\frac{1}{N_0}\sum_{m=1}^{N}\mathbf{1}\{m\in I\}\,\delta_{\mathbf{y}_m}(A)\right]
&=
\frac{1}{N_0}\sum_{m=1}^{N}\frac{N_0}{N}\,\delta_{\mathbf{y}_m}(A),\\
&=
\frac{1}{N}\sum_{m=1}^{N}\delta_{\mathbf{y}_m}(A),\\
&=
\widehat{F}_N(A).
\end{align*}

\emph{Remarks.} (i) The result holds for any distribution of \(N_0\) provided that, conditional on \(N_0\), the subset \(I\) is uniformly distributed. (ii) The Dirichlet concentration (or the CV-derived \(\kappa\)) may depend on \(N_0\); symmetry ensures \(\mathbb{E}[w_j\,|\,I,N_0]=1/N_0\), so centering is unaffected. 

\end{document}